\pdfoutput=1

\documentclass[11pt]{article}

\usepackage[]{EMNLP2023}

\usepackage[T1]{fontenc}    
\usepackage[utf8]{inputenc} 
\usepackage{times}
\usepackage{latexsym}
\usepackage{amsmath, amsfonts}
\usepackage{microtype}
\usepackage{inconsolata}

\usepackage{booktabs}
\usepackage{multirow}
\usepackage{makecell}
\usepackage{tabularx}
\usepackage{colortbl}
\usepackage{xcolor}
\usepackage[table]{xcolor} 

\usepackage{graphicx}
\usepackage{tikz}
\usetikzlibrary{arrows.meta,positioning,shapes,fit,backgrounds,calc,shadows.blur}

\usepackage{algorithm}
\usepackage{algorithmic}

\usepackage{hyperref}
\usepackage{url}

\definecolor{RoleClaim}{RGB}{255,244,179}    
\definecolor{RoleSupport}{RGB}{198,227,255}  
\definecolor{RoleOppose}{RGB}{255,204,204}   

\newcommand{\ent}[1]{\textbf{#1}}
\newcommand{\argrole}[2]{\colorbox{#1}{\strut #2}}

\usepackage{array}
\newcolumntype{Y}{>{\raggedright\arraybackslash}X}

\title{\texttt{SynClaimEval}: A Framework for Evaluating the Utility of Synthetic Data in Long-Context Claim Verification}

\author{
  Mohamed Elaraby\thanks{ Work done during an internship with Zillow.} \\
  University of Pittsburgh \\
  \texttt{mse30@pitt.edu} 
  \And
  Jyoti Prakash Maheswari \\
  Zillow Inc. \\
  \texttt{jyotip@zillowgroup.com}
}

\begin{document}
\maketitle
\begin{abstract}
Large Language Models (LLMs) with extended context windows promise direct reasoning over long documents, reducing the need for chunking or retrieval. Constructing annotated resources for training and evaluation, however, remains costly. Synthetic data offers a scalable alternative, and we introduce \textbf{\texttt{SynClaimEval}}, a framework for evaluating synthetic data utility in \emph{long-context claim verification}—a task central to hallucination detection and fact-checking. Our framework examines three dimensions: (i) \textit{input characteristics}, by varying context length and testing generalization to out-of-domain benchmarks; (ii) \textit{synthesis logic}, by controlling claim complexity and error type variation; and (iii) \textit{explanation quality}, measuring the degree to which model explanations provide evidence consistent with predictions. Experiments across benchmarks show that long-context synthesis can improve verification in base instruction-tuned models, particularly when augmenting existing human-written datasets. Moreover, synthesis enhances explanation quality, even when verification scores don't improve, underscoring its potential to strengthen both performance and explainability.   
\end{abstract}


\section{Introduction}

\begin{figure*}[t!]
    \centering
    \includegraphics[width=\textwidth]{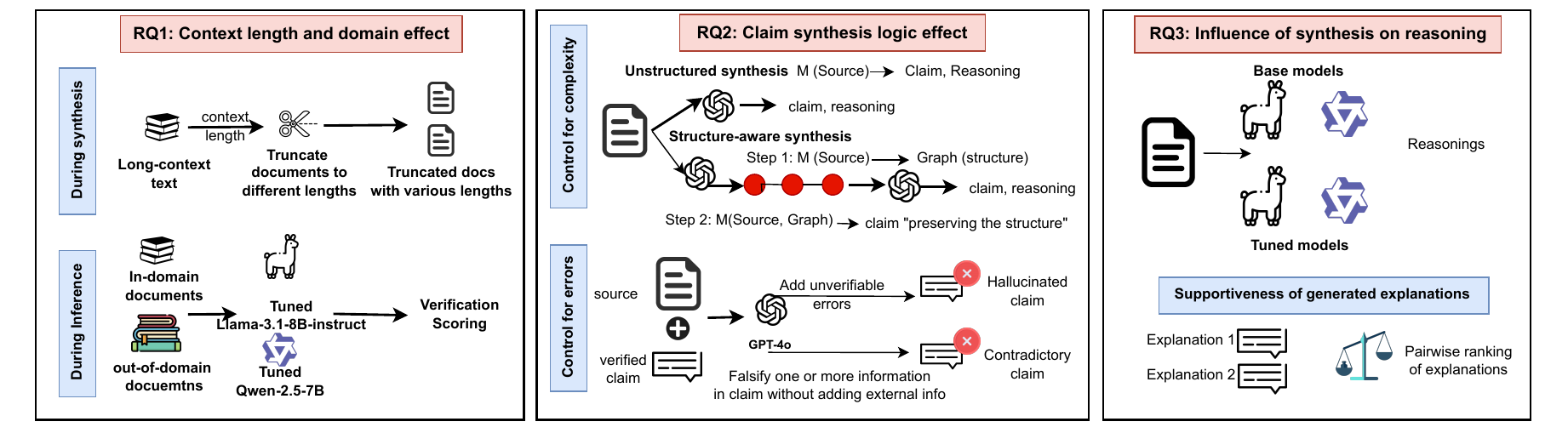}
    \caption{\centering Overview of the \textbf{\texttt{SynClaimEval}} pipeline. The framework is designed to evaluate synthetic data along three dimensions: (1) \textit{context length and domain effects}, (2) \textit{claim generation logic}, and (3) \textit{explanation quality}.}
    \label{fig:eval_pipeline}
\end{figure*}

Extending the context window of large language models (LLMs) to process thousands and millions of tokens is a promising step toward building systems capable of comprehending long, complex documents without relying on aggressive chunking or retrieval-based pipelines \cite{liu2025comprehensive}. However, constructing datasets for both fine-tuning and evaluating long-context LLMs remains labor-intensive and costly, limiting scalability. 
Synthetic datasets have emerged as a promising alternative to manual annotation, enabling large-scale, low-cost generation of training and evaluation data \cite{viswanathan-etal-2025-synthetic}. Yet, in the long-context setting, empirical findings remain mixed: some studies report diminished or even negative effects from synthetic long-context training \cite{gao2024train}, while others demonstrate substantial gains over weak long-context baselines \cite{pham2025clipper}. These discrepancies highlight the need for a systematic evaluation of synthetic data’s utility in improving long-context reasoning. In this work, we focus on \textbf{evaluating long-context synthesis for long-context claim verification task}.

We pose the following research questions (RQs), addressing both verification performance and explanation quality. 
\textbf{RQ1: How does synthetic long-context training data affect downstream claim?} We study this question along two dimensions: (\textit{i}) the effect of context length on verification accuracy, and (\textit{ii}) the impact of the source domain of the synthetic data on out-of-domain verification benchmarks. 
\textbf{RQ2: How does synthesis logic affect downstream claim verification? } 
We study this by varying \textit{error types} in unverifiable claims and \textit{claim complexity} in verifiable ones.  
\textbf{RQ3: Does synthetic training improve the quality of model-generated explanations?} 
We examine whether synthetic tuning improves explanation quality by encouraging rationales that more consistently cite relevant evidence from the input context.

 We introduce \textbf{\texttt{SynClaimEval}}, an evaluation framework for systematically evaluating the utility of synthetic data in long-context claim verification across the dimensions outlined in our research questions. Figure~\ref{fig:eval_pipeline} provides an overview of the framework.  
For \textbf{RQ1}, we vary training context length by truncating source articles, while keeping evaluation benchmarks untruncated as reference, and test both within-domain and out-of-domain settings to assess generalization.  
For \textbf{RQ2}, we manipulate the logic of synthesis along two dimensions: \textit{complexity}, by conditioning on structured representations that induce multi-hop reasoning, and \textit{error type}, by contrasting hallucinated (unverifiable) claims with contradictory ones.  
For \textbf{RQ3}, we evaluate explanation quality through pairwise ranking, asking whether rationales generated under different synthesis strategies offer more support to the same predicted label.

Our study yields five key insights: (i) long-context synthesis enables base instruction-following models to narrow the gap with stronger models, though gains are not always consistent; (ii) extending training contexts improves verification performance; (iii) balancing contradictory and unverifiable (hallucinated) errors yields larger improvements than relying solely on unverifiable errors; (iv) structured synthesis (e.g., multi-hop reasoning) improves performance and generalizes more effectively than unstructured approaches; and (v) although verification gains are modest, synthesis consistently improves explanation quality, independent of verification accuracy improvements.



\section{Related Work}

\noindent \textbf{Long-context Claim Verification} 
Early work on claim verification largely relied on natural language inference (NLI) models such as \texttt{BERT} \cite{devlin-etal-2019-bert}, \texttt{RoBERTa} \cite{liu2019roberta}, and \texttt{DeBERTa} \cite{hedeberta}, which were limited to short contexts \cite{kryscinski-etal-2020-evaluating}. To adapt these models for longer inputs, prior approaches typically truncated documents \cite{zha-etal-2023-alignscore, zhang-etal-2024-fine} or used retrieval-based strategies \cite{bishop2024longdocfactscore}. More recently, advances in position interpolation and extrapolation have enabled LLMs to process extended contexts directly \cite{presstrain, pengyarn}, motivating the development of long-context verification benchmarks. For example, \citet{zhao-etal-2024-findver} introduced a financial benchmark where even state-of-the-art models (e.g., Claude-3.5) fall far behind human experts, while \citet{karpinska-etal-2024-one} proposed a benchmark for verifying claims across fictional books. \textit{In this work, we address long-context claim verification from a broader perspective: rather than targeting a specific domain, we study how synthetic data derived from public benchmarks can serve as effective tuning resources that generalize across diverse long-context settings.}

\noindent \textbf{Synthetic Data in Claim Verification} 
Claim verification can be framed as an entailment task, where most widely used datasets are short-context and human-authored across diverse domains \cite{bowman2015large, williams-etal-2018-broad}. In contrast, human-written long-context resources are scarce and often domain-specific, such as legal contracts 
Synthetic data has shown promise in extending verification tasks: for short contexts, \citet{tang-etal-2024-minicheck} proposed two synthesis pipelines that augmented existing NLI benchmarks, yielding performance comparable to \texttt{GPT-4o}. Building on this, \citet{lei-etal-2025-factcg} demonstrated that generating claims from context graphs improves over direct prompting, especially for multi-hop reasoning. Results in long-context settings, however, remain mixed. Some studies suggest that short-context synthesis is sufficient for generalization to longer documents \cite{gao2024train, bai-etal-2024-longalign}, while others show that in claim verification—particularly narrative domains—long-context synthesis, often from compressed document representations, yields stronger results \cite{pham2025clipper}. 
\textit{In this work, we systematically explore long-context claim synthesis with a strong LLM, evaluating unexplored dimensions such as the effect of error types, varying claim complexity, cross-domain generalization, and the impact of synthesis on explanation quality.}

\begin{table*}[t]
\small
\setlength{\tabcolsep}{6pt}
\renewcommand{\arraystretch}{1.22}
\centering
\begin{tabularx}{\textwidth}{l Y}
\toprule
\textbf{Row} & \textbf{Content (verifiable-only examples)} \\ 
\midrule
\makecell[l]{\textbf{Summary}\\(sinppet)} 
&
\textit{The report examines the \ent{Senators' Official Personnel and Office Expense Account (SOPOEA)}, which funds staff salaries, travel, supplies, and other office costs. \argrole{RoleClaim}{\ent{The largest expenditure category} is \ent{personnel compensation}, which accounts for approximately} \argrole{RoleClaim}{\ent{90\%} of total SOPOEA spending}. 
Across selected fiscal years (\ent{2007}, \ent{2008}, \ent{2011}, \ent{2012}), \argrole{RoleSupport}{spending categories are largely consistent and overall trends remain relatively stable}. \argrole{RoleOppose}{There is still variation across spending categories and overall funding levels have decreased} \argrole{RoleOppose}{or remained flat in recent years}. 
The allocation formula depends on \ent{population} and \ent{distance from Washington, DC}, and the \ent{Senate Appropriations Committee} periodically adjusts SOPOEA limits to emphasize transparency and prudent spending.}
\\[2pt]
\midrule

\textbf{Unstructured claim} 
&
\textbf{Claim:} \emph{\ent{Personnel compensation} accounts for approximately \ent{90\%} of total \ent{SOPOEA} spending.} \\[2pt]
\midrule

\makecell[l]{\textbf{Context-graph}\\(entities \& path)} 
&
\textbf{3-Hop Path:} \(\ent{SOPOEA} \xrightarrow{\texttt{has\_category}} \ent{personnel\_compensation} \xrightarrow{\texttt{accounts\_for}} \ent{90\%} \xrightarrow{\texttt{implies}} \ent{largest\_category}\) \\
& 
\textbf{Claim:} \emph{Within \ent{SOPOEA}, \ent{personnel compensation} constitutes about \ent{90\%} of total spending making it the \ent{largest category}.} \\[2pt]
\midrule

\makecell[l]{\textbf{Argument-graph}\\(roles \& polarity)} 
&
\textbf{Chain:} 
\argrole{RoleClaim}{Claim} \(\leftarrow\) 
\argrole{RoleOppose}{Premise (opposes)} \\[-2pt]
& 
\textbf{Generated Claim:} \argrole{RoleClaim}{Personnel compensation consistently represents the largest expenditure category in SOPOEA} \argrole{RoleClaim}{spending, accounting for approximately \ent{90\%} of total expenditures,} \argrole{RoleOppose}{despite variations in other spending categories and overall funding levels.} \\
\bottomrule
\end{tabularx}
\caption{\centering Verifiable claims examples. 
Entities are \textbf{bolded}. 
Arguments are highlighted: 
\argrole{RoleClaim}{\textbf{Claim}}, 
\argrole{RoleSupport}{\textbf{Supporting Premise}}, 
\argrole{RoleOppose}{\textbf{Opposing Premise}}.} 
\label{tab:verifiable_examples_sopoea}
\end{table*}

\section{\texttt{SynClaimEval}}
In this section, we describe the components of our evaluation framework.  

\subsection{Preparing Claim Sources}
\label{subsec:prepare_sources}

\noindent \textbf{Document Truncation}  
For \textbf{RQ1}, we examine how context length affects continual supervised fine-tuning (SFT) with synthetic claims. To simulate different source configurations, each document is truncated to a maximum length $T \in \{4{,}096, 8{,}192, 16{,}384\}$ tokens. This design allows us to directly compare models trained on shorter versus longer contexts under identical evaluation conditions, while preserving the integrity of the source.  

\noindent \textbf{Compression-based Claim Synthesis.}  
Following \texttt{CLIPPER} \cite{pham2025clipper}, we synthesize claims from compressed document representations (summaries), which produce less noisy and more cost-effective claims than generating directly from full long-context inputs.  
  We leverage \texttt{GPT-4o} to generate a summary of no more than $1{,}000$ words by instructing the model to produce a concise version of the truncated document. This compressed summary then serves as the source for claim synthesis. To account for domain-specific characteristics in our synthesis sources, we design a dedicated summarization prompt for each domain type\footnote{Summarization prompts are provided in Appendix~\ref{app:summ_prompts}}.

\subsection{Claim Synthesis Strategies \footnote{We use \texttt{GPT-4o} as the synthesizer. All prompts are in \ref{app:generation_prompts}}} 
\label{subsec:claim_synth_strateg}

We design a synthetic data generation pipeline that produces claims varying along two key axes. 
First, we control \textit{complexity}: unstructured claims are generated directly from the source text (summaries), while structured claims require multi-hop reasoning either across entities or across discourse/argument units in the context. Second, we vary the \textit{error type}, generating both unverifiable claims that introduce hallucinated content and contradictory claims that embed factual errors. 
Algorithm~\ref{alg:synthesis} outlines the generic synthesis framework. 

\begin{algorithm}[h]
\caption{Generic Claim Synthesis Framework}
\label{alg:synthesis}
\begin{algorithmic}[1]
\STATE \textbf{Input:} (Document $D$ , summary $S$)
\STATE Extract structured representation $I \leftarrow f_{\text{struct}}(S)$
\IF{Unstructured mode}
  \STATE $I \leftarrow S$
\ELSE
  \STATE $I \leftarrow f_{\text{struct}}(S)$ \\extract structure from text
\ENDIF
\STATE Generate verifiable claims: $C^{+}\leftarrow f_{\text{claim}}(I,S)$
\STATE Generate unverifiable variants: $C^{u}\leftarrow f_{\text{unverif}}(I,S, C^{+})$
\STATE Generate contradictory variants: $C^{c}\leftarrow f_{\text{contrad}}(I,S, C^{+})$
\STATE \textbf{Output:} Synthetic set $\mathcal{S} = \{(D,C^{+}),(D,C^{u}),(D,C^{c})\}$
\end{algorithmic}
\end{algorithm}

\noindent \textbf{Unstructured Synthesis.}
 We directly prompt the LLM with $(S,D)$ to generate verifiable claims $
C^{+}\leftarrow f_{\text{claim}}(S,D).$
To generate error variants, we obtain unverifiable claims by $C^{u}\leftarrow f_{\text{unverif}}( C^{+},D),$
which takes the verifiable claim $C^{+}$ and inserts plausible but unsupported facts that are not grounded in $D$. Contradictory claims are obtained by:
$C^{c}\leftarrow f_{\text{contrad}}(C^{+}, D),$
where $f_{\text{contrad}}$ applies common error transformations obtained from the error taxonomy in \cite{mishrafine,devaraj-etal-2022-evaluating,pagnoni-etal-2021-understanding}. Namely we include negation, entity errors, or discourse polarity reversal \footnote{Appendix \ref{app:err_types} includes error types definitions and examples}. Table \ref{tab:verifiable_examples_sopoea}, second row, shows an example of generated unstructured verifiable claim synthesized from the summary. 

\noindent \textbf{Context-graph Synthesis.}
Many claims in long contexts require reasoning over entity relations spanning multiple document segments. To simulate this, we follow the method in \cite{lei-etal-2025-factcg} by constructing a \emph{context graph} $G = (V,E)$ by prompting an LLM to extract entity–relation triplets from summary $S$. We normalize triplets and form non-branching connected components. From $G$, we sample multi-hop paths $\pi_{\text{entity}}$ of length up to $k=3$ \footnote{More hops do not yield further improvement}. Verifiable claims $C^{+}$ are generated by
$
f_{\text{claim}} : (S,\pi_{\text{entity}}) \mapsto C^{+}$. Unverifiable claims $C^{u}$ are obtained by inserting unsupported relations, while contradictory claims $C^{c}$ are created by corrupting existing edges (e.g., reversing relation types). Table \ref{tab:verifiable_examples_sopoea}, third row, shows an example of an extracted $3$-hop path from the entities and how they are aggregated into one single claim.
\paragraph{Argument-graph Synthesis.}
Building on prior work in claim verification that leverages composite evidence roles \cite{habernal2018semeval}, and recent advances in argumentative LLMs that demonstrate improvements in the explainability of verifiable claims \cite{freedman2025argumentative}, we extend these insights to structured synthesis for long-context verification.  
 We introduce a synthesis strategy that leverages \emph{argument graphs} to capture multi-hop argumentative reasoning. 
In this formulation, we construct an argument graph $A=(V,E)$, where nodes $V$ represent argumentative units (claims or premises) and edges $E$ encode polarity relations (\textit{supports}, \textit{opposes}). 
Argument roles are extracted from $S$ using an LLM-based argument-mining prompt. 
From $A$, we then sample coherent chains $\pi_{\text{arg}}$ that connect a central claim to its supporting and/or opposing premises. 
This design simulates claim synthesis that relies on reasoning across multiple argumentative evidence, rather than purely entity-based links, exposing models to more discourse-level verification challenges. The remainder of the synthesis pipeline mirrors the context-graph setup: given an extracted chain, we first generate a verifiable claim, which is then perturbed to produce its unverifiable and contradictory variants.  
Table \ref{tab:verifiable_examples_sopoea}, final row, shows an example of a generated claim based on two rhetorical roles where the premise opposes the claim. The synthesized claim is controlled to capture the relation between them, yielding more complex claims at the sentence level.

\begin{table*}[t]
\centering
\label{tab:synthetic_4k_16k_stats}
\resizebox{\textwidth}{!}{%
\begin{tabular}{lcrrrll}
\toprule
\textbf{Variant} & \textbf{Truncation} & \textbf{Total Claims} & \textbf{Verified (n)} & \textbf{Unverified (n)} & \textbf{Claim len (min/mean/max)} & \textbf{Reasoning len (min/mean/max)} \\
\midrule
\multirow{2}{*}{Unstructured}
 & 4k  & 14{,}074 & 2{,}815 & 11{,}259 & 6 / 23.17 / 187 & 13 / 31.83 / 100 \\
 & 8k& 14{,}072& 2{,}815& 11{,}257& 4 / 20.70 / 90&11 / 29.77 / 80\\
 & 16k & 14{,}072 & 2{,}815 & 11{,}257 & 5 / 23.41 / 102 & 12 / 31.78 / 87 \\
\addlinespace
\multirow{2}{*}{Context-graph Synthesis }
 & 4k  & 8{,}403 & 2{,}793 & 5{,}610 & 7 / 32.46 / 124 & 17 / 46.85 / 99 \\
 & 8k& 7{,}882& 2{,}420& 5{,}462& 7 / 32.63 / 111&16 / 43.65 / 118\\
 & 16k & 8{,}421 & 2{,}803 & 5{,}618 & 7 / 32.71 / 148 & 16 / 46.82 / 110 \\
\addlinespace
\multirow{2}{*}{Argument-graph Synthesis}
 & 4k  & 7{,}977 & 2{,}672 & 5{,}305 & 6 / 44.95 / 259 & 16 / 65.86 / 198 \\
 & 8k& 6{,}156& 2{,}048& 4{,}108& 5 / 44.06 / 208&10 / 58.64 / 140\\
 & 16k & 7{,}970 & 2{,}687 & 5{,}283 & 6 / 45.04 / 473 & 12 / 65.64 / 221 \\
\bottomrule
\end{tabular}
}%
\caption{\centering Claim distribution and claim length statistics (in words) across all training synthesis strategies. 
}
\label{tab:synth_stats}
\end{table*}

\subsection{Evaluating Explanations (RQ3)}
\label{subsec:explain_eval}
We assess \emph{justification strength}, i.e., how well an explanation provides valid and sufficient evidence from the context to support the predicted label. 
Following \citet{elaraby-etal-2024-persuasiveness}, we frame this as a pairwise ranking task, comparing explanations from different models or tuning strategies against the untuned baseline. 
Given two explanations ($e_i$, $e_j$) for the same claim and predicted label $l \in \{True, False\}$, we use \texttt{GPT-4o} to judge which better supports the decision. 
Each explanation earns $1$ point per win and $0.5$ per tie:
\[
s_i = \sum_{\substack{j=1 \\ j \neq i}}^M \mathbb{I}[e_i > e_j] 
      + 0.5 \sum_{\substack{j=1 \\ j \neq i}}^M \mathbb{I}[e_i = e_j],
\]
where $\mathbb{I}$ denotes the judge’s preference. 
We report average ranking scores across benchmarks.

\section{Datasets}

\subsection{Synthetic Sources}
We construct our synthetic data from widely used, publicly available long-context benchmarks: \texttt{PubMed} \cite{cohan-etal-2018-discourse}, \texttt{GovReports} \cite{huang-etal-2021-efficient}, \texttt{MeetingBank} \cite{hu-etal-2023-meetingbank}, and \texttt{SQuality} \cite{wang-etal-2022-squality}. 
These datasets were selected to provide a diverse set of domains, enabling us to evaluate the utility of synthesis across varied and openly accessible benchmarks.
We uniformly sampled $900$ documents from the four datasets, ensuring no overlap with those included in our test benchmarks. Of these, $600$ \footnote{Comparable training source sizes are also used in \cite{pham2025clipper}} serve as training sources, while the remaining $300$ are reserved to construct an in-domain synthetic test set.
  
\noindent \textbf{Filtration and Truncation.}  
For both training and testing sources, we exclude documents  $< 1024$ tokens. We then apply the pipeline in \S\ref{subsec:prepare_sources}. Truncation is applied only to training sources to simulate the effect of context length on benchmarks, while test documents are preserved in their full length. 

\noindent \textbf{Obtaining Synthetic Training.}  
We apply both unstructured and structured synthesis strategies as described in \S\ref{subsec:claim_synth_strateg}. For each strategy, we sample an equal number of \emph{verified} and \emph{unverified} claims to ensure balanced supervision.  
To study the impact of error type, we construct two parallel training sets for each synthesis strategy:  
(1) an \emph{unverified-only} set, where all negative pairs correspond to unverified errors, and  
(2) a \emph{diverse-error} set, where negative pairs are evenly split between unverified errors (hallucinations) and contradictory errors (balanced across contradiction types).  
This design allows us to isolate the effect of different error distributions on model training.
Table~\ref{tab:synth_stats} summarizes statistics for the synthetic training datasets across synthesis strategies. Unstructured synthesis yields the largest number of claims, since generating contradictory variants naturally increases error diversity. Truncation has only a minor effect on claim counts and lengths, reducing the risk of confounds when analyzing truncation during fine-tuning. In contrast, structured synthesis produces longer claims and reasoning spans, reflecting our design choice to encourage more complex, multi-faceted examples.

\noindent \textbf{Quality of Generated Claims \footnote{Automatic quality evaluation of synthetic claims is in \ref{app:auto_claim_eval} and of synthetic explanations in \ref{app:exp_quality} }} 
We employed three annotators to validate the quality of synthetic claims, ensuring no confounding errors from the synthesis process.  
From the $4k$ unstructured-context set (avoiding longer contexts for efficiency), we sampled $540$ claims evenly across types ($180$ verifiable, $180$ unverifiable, $180$ contradictory) \footnote{Annotators only disagreed on $14$ samples out of the $540$ $\text{IAA}=0.991\%$}.  
Annotators checked each claim’s assigned label against its source context, yielding agreement rates of $97.22\%$, $97.77\%$, and $99.16\%$ for verifiable, unverifiable, and contradictory claims, respectively—demonstrating the high purity of our synthetic pipeline.


\subsection{Evaluation Benchmarks}

We evaluate fine-tuning on both synthetic test sets from \texttt{SynClaimEval}, aligned with the training distributions, and on publicly available long-document benchmarks with claim- or statement-level support annotations.

\noindent \textbf{\texttt{SynClaimEval}}  
We applied the unstructured synthesis pipeline to $300$ source documents that were not part of training or any publicly available benchmark. We deliberately avoided constructing a structured synthesis test set in order to assess whether models trained on structured claims can generalize to unstructured settings, where the error distribution differs. In total, we generated $2{,}500$ claims evenly distributed across the labels: verified, unverified, negation, entity error, and discourse error.  

\noindent \textbf{\texttt{UniSummEval}} \footnote{\url{https://github.com/DISL-Lab/UniSumEval-v1.0}} \cite{wang-etal-2022-squality}%
 is a summarization evaluation benchmark constructed from widely used long-context datasets: \texttt{PubMed}, \texttt{GovReports} , \texttt{MeetingBank}, \texttt{SQuality}, and \texttt{MediaSumm}. 
 \begin{table}[h!]
\small
\centering
\resizebox{\columnwidth}{!}{
\begin{tabular}{lcccc}
\toprule
\textbf{Benchmark} & \textbf{\# Pos.} & \textbf{\# Neg.} & \textbf{Claim len.} & \textbf{Context len.} \\
\midrule
\texttt{SynClaimEval (Test)} & $500$ & $2000$ & $6/22/76$ & $54/4921/31923$ \\ 
\texttt{UniSummEval}       & $4897$ & $402$  & $2/23/97$ & $293/3903/10462$ \\ 
\texttt{FinDver}           & $350$  & $350$  & $11/38/87$ & $4160/39866/69724$ \\ 
\bottomrule
\end{tabular}
}
\caption{\centering Statistics of included test benchmarks. }
\label{tab:dataset_stats}
\end{table}
 Each summary sentence is annotated with a binary label indicating whether it is fully supported by the input context. The benchmark covers both short- and long-context documents; in this work, we focus exclusively on the \texttt{"long"} subset, yielding $5{,}299$ sentence–document pairs. Our motivation for using \texttt{UniSummEval} is to evaluate models tuned on \texttt{SynClaimEval} against a large, multi-domain benchmark that shares the same document characteristics as training, but differs in downstream task framing.

\noindent \textbf{\texttt{FinDVer}}    \footnote{\url{https://github.com/yilunzhao/FinDVer}} \cite{zhao-etal-2024-findver}%
 is a long-context financial document benchmark in which claim verification requires reasoning across multiple sections of a document. Verifying these claims often entails identifying and correctly interpreting the relevant evidence within the text. We use the \textit{test-mini} split, which contains $700$ long financial reports paired with annotated claims and their corresponding reasoning. Our motivation for including \texttt{FinDVer} is to test \texttt{SynClaimEval} on more complex and out-of-domain long-context benchmarks where long context LLMs are known to struggle to verify the claims against them. 


Table~\ref{tab:dataset_stats} summarizes the overall statistics of the included test beds. For our in-domain synthetic test set, the average claim length is comparable to that of the \texttt{UniSummEval} benchmark, which is expected given the shared source domains used for synthesis. Among the public benchmarks, \texttt{FinDver} contains the longest documents on average, a characteristic that is reflected in its relatively longer claims. In contrast, \texttt{UniSummEval} shows a strong skew toward positive claims, which is unsurprising since its claims are derived from sentences in generated summaries—a task where LLMs have been shown to perform strongly \cite{changbooookscore}.

\begin{figure}[t]
  \centering
\includegraphics[width=0.49\textwidth]{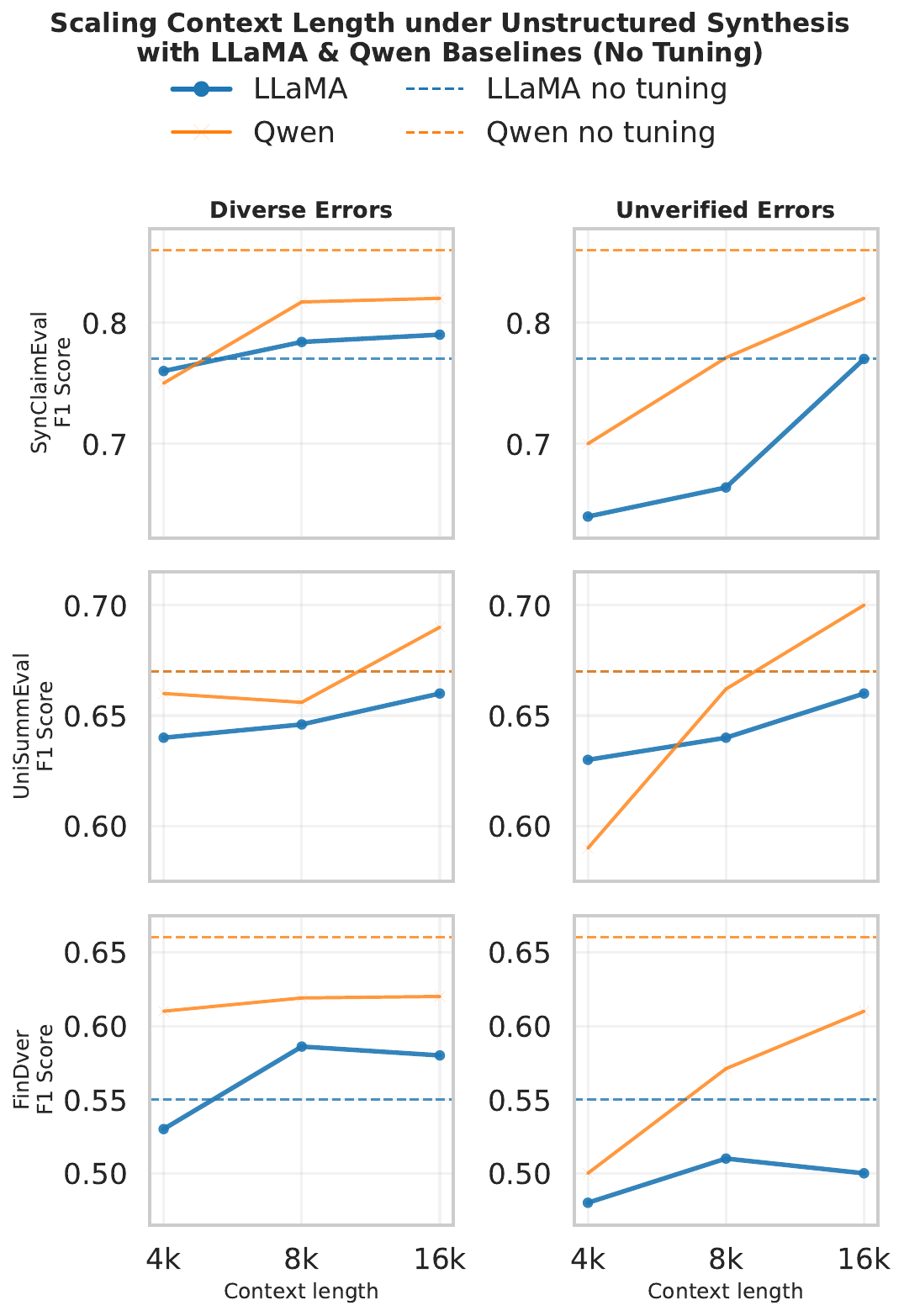}
  \caption{\centering Context length effect on scoring}
  \label{fig:context_len_trend}
\end{figure}
\section{Experimental Setup} 

\subsection{Models and Prompting} 
We evaluate long-context LLMs with $>$120k token capacity, including proprietary (\texttt{GPT-4o}, \texttt{GPT-4o-mini}) and open-weight (\texttt{LLaMA-3.1-8B-Instruct} (\texttt{LLaMa}) \cite{grattafiori2024llama}, \texttt{Qwen-2.5-7B-Instruct} (\texttt{Qwen}) \cite{yang2024qwen2}, interpolated linearly from 32k$\to$128k).  
For both inference and tuning, we use the \texttt{BeSpoke} prompt from \texttt{MiniCheck}~\cite{tang-etal-2024-minicheck}, which requires a binary decision (\texttt{yes}/\texttt{no}) and a free-text explanation; decoding temperature is fixed to $0$.  

\subsection{Continual Fine-tuning} 
Continual SFT is performed with \texttt{QLoRA} \cite{dettmers2023qlora} (4-bit, rank=16, $\alpha=32$), training each model for two epochs.\footnote{Larger ranks/$\alpha$ offered no gains.}  
As a baseline, we fine-tune on $16892$ human-written samples from \texttt{ANLI} \cite{nie-etal-2020-adversarial}, following prior work showing short-context tuning may transfer to long contexts \cite{grattafiori2024llama, gao2024train} and to measure utility of synthetic long context datasets against human written short ones. For synthetic tuning, we construct 4k balanced pairs (2k verified, 2k unverified), split 85/15 into train/validation. We also evaluate hybrid settings that augment \texttt{ANLI} with synthetic data, extending strategies effective in short-context verification \cite{tang-etal-2024-minicheck}.


\section{Results and Analysis}


\subsection{RQ1: Context Length and Domain Generalization}

\noindent \textbf{Context Length.}  We first isolate the effect of input length by truncating source documents, holding synthesis complexity fixed through the unstructured variant.  
Figure~\ref{fig:context_len_trend} shows that for both \texttt{LLaMA} and \texttt{Qwen}, expanding the context window  consistently improves verification performance. 
This pattern is consistent with prior findings \cite{pham2025clipper}, which similarly reported that longer contexts yield stronger supervision for claim verification.   
In subsequent experiments, we therefore fix the training context length at 16k to focus on the effect of synthesis complexity (RQ2).  

\noindent\textbf{Generalization}  
Figure \ref{fig:context_len_trend} On in-domain and near-domain tests (\texttt{SynClaimEval}, \texttt{UniSummEval}), \texttt{LLaMA} shows clear gains at 16k over its non-tuned baseline, whereas \texttt{Qwen} underperforms its already strong baseline, which outperforms \texttt{LLaMA} across all benchmarks.  
This suggests that unstructured synthesis can help weaker models narrow the gap but provides limited benefit for models that already perform well. We further investigate whether more complex claims improve generalization in RQ2.

\begin{figure}[t]
  \centering
\includegraphics[width=0.5\textwidth]{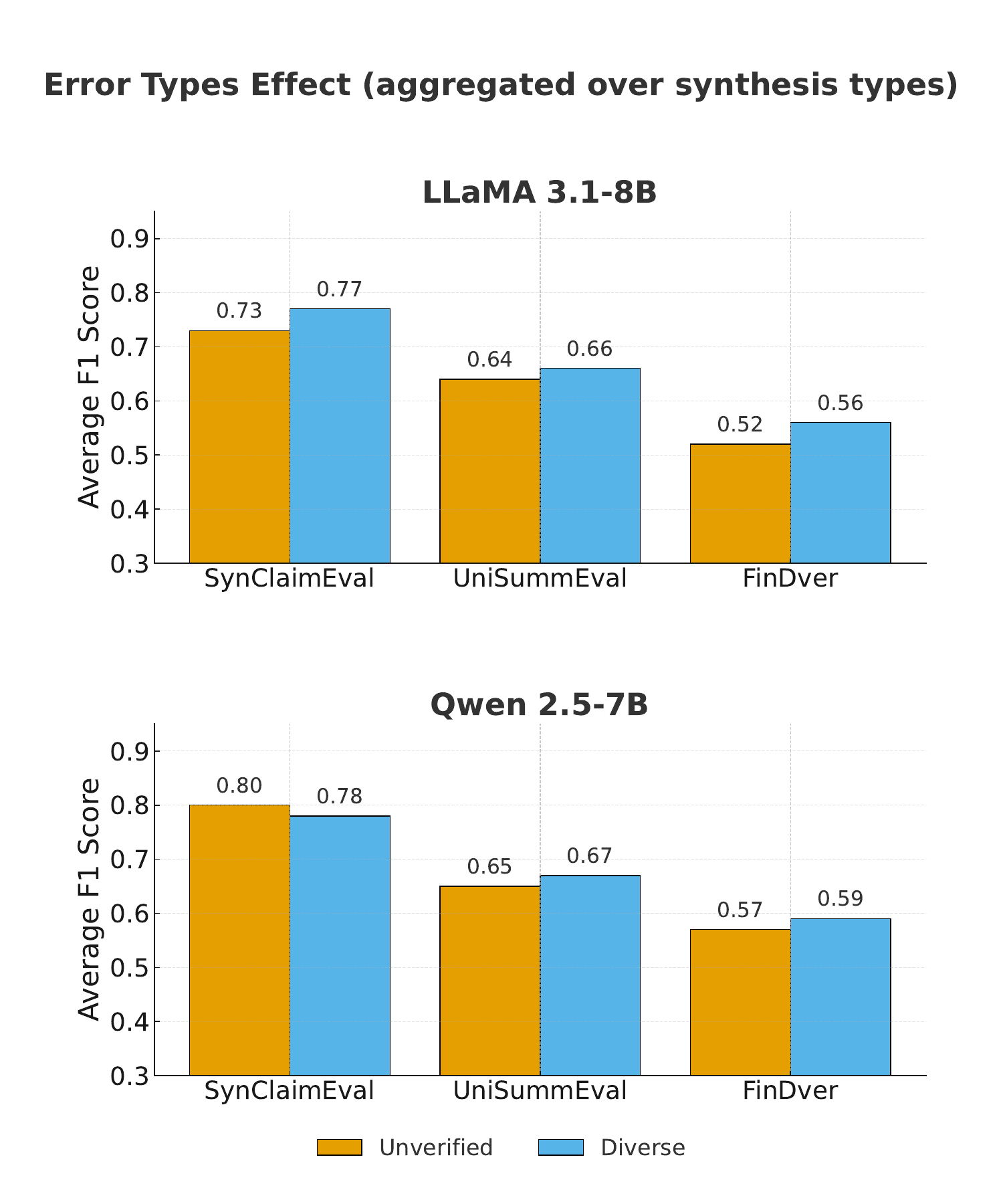}
  \caption{\centering Error types effect}
  \label{fig:error_types}
\end{figure}

\begin{figure*}[t]
  \centering
  \includegraphics[width=0.95\textwidth]{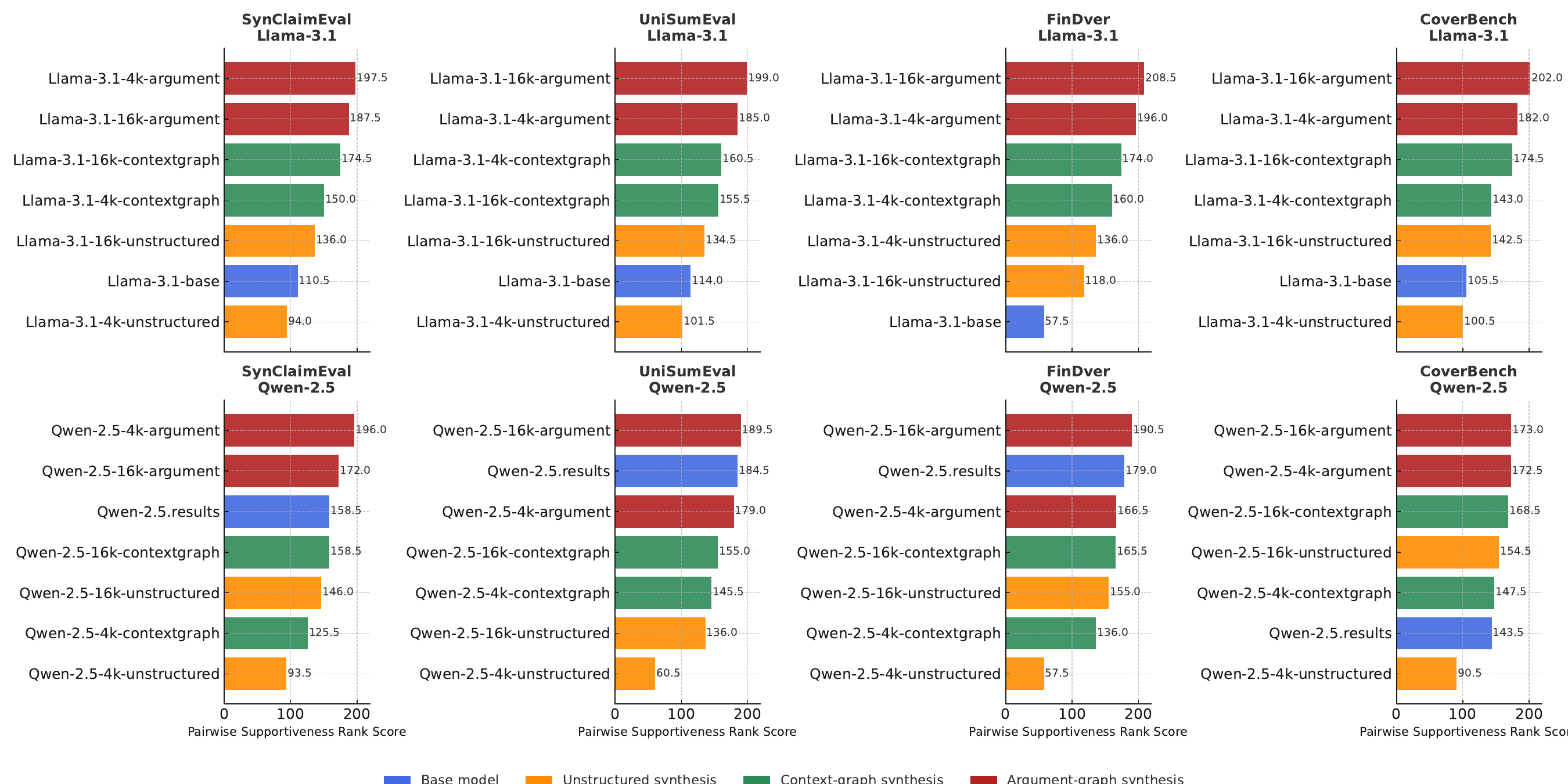}
  \caption{\centering Pairwise supportiveness ranking of explanations across benchmarks. Colors denote synthesis type (Base, Unstructured, Context-graph, Argument-graph). Higher scores indicate stronger judged quality.}
  \label{fig:expl_ranking_composite}
\end{figure*}

\begin{table}[ht!]
\small
\centering
\resizebox{\columnwidth}{!}{
\begin{tabular}{l|c|c|c}
\toprule
\textbf{Model / Setting} &
\textbf{\texttt{SynClaimEval} F1} &
\textbf{UniSummEval F1} &
\textbf{FinDver F1} \\
\midrule
\multicolumn{4}{l}{\textit{\textbf{Baselines (Proprietary)}}} \\
\textit{GPT-4o}        & \textbf{0.97} & \textbf{0.71} & \textbf{0.81} \\
\textit{GPT-4o-mini}   & 0.93          & \textbf{0.71} & 0.74 \\
\midrule
\multicolumn{4}{l}{\textit{\textbf{Baselines (Open-weight)}}} \\
\textit{LLaMA-3.1-8B}  & 0.77 & 0.67 & 0.55 \\
\textit{Qwen-2.5-7B}   & 0.86 & 0.67 & 0.66 \\
\midrule

\multicolumn{4}{l}{\textbf{Unstructured synthesis}} \\
LLaMA-3.1-8B           & 0.77 & 0.66 & 0.50 \\
\rowcolor{gray!10}
LLaMA-3.1-8B           & \underline{\textit{0.79}} & 0.66 & \underline{\textit{0.58}} \\
Qwen-2.5-7B            & 0.82 & \underline{0.70} & 0.61 \\
\rowcolor{gray!10}
Qwen-2.5-7B            & 0.82 & \underline{0.69} & 0.62 \\
\midrule

\multicolumn{4}{l}{\textbf{Context-graph (structured)}} \\
LLaMA-3.1-8B           & \underline{\textit{0.79}} & \underline{\textit{0.69}} & 0.52 \\
\rowcolor{gray!10}
LLaMA-3.1-8B           & \underline{0.78} & \underline{0.68} & \underline{0.57} \\
Qwen-2.5-7B            & 0.82 & \underline{0.70} & 0.61 \\
\rowcolor{gray!10}
Qwen-2.5-7B            & 0.81 & \underline{0.70} & 0.62 \\
\midrule

\multicolumn{4}{l}{\textbf{Argument-graph (structured)}} \\
LLaMA-3.1-8B           & \underline{\textit{0.82}} & 0.62 & \underline{\textit{0.58}} \\
\rowcolor{gray!10}
LLaMA-3.1-8B           & \underline{0.79} & 0.66 & \underline{0.57} \\
Qwen-2.5-7B            & 0.79 & \underline{0.69} & 0.60 \\
\rowcolor{gray!10}
Qwen-2.5-7B            & 0.79 & \underline{0.70} & 0.60 \\
\bottomrule

\multicolumn{4}{l}{\textit{\textbf{Blended Synthetic dataset with and without ANLI}}}\\
\rowcolor{green!10}
\textit{LLaMA-3.1-8B}  & 0.72 & 0.65 & \underline{0.61} \\
\rowcolor{gray!10}
LLaMA-3.1-8B           & \underline{0.81} & 0.64 & \underline{0.63} \\
\rowcolor{blue!10}
LLaMA-3.1-8B           & \underline{\textit{0.82}} & 0.64 & \underline{\textit{0.65}} \\ 
\bottomrule
\end{tabular}
}
\caption{\centering Performance across benchmarks in F1.  
\underline{Underline} = fine-tuned improvements; 
\textit{Italics} = best among \emph{LLaMA} rows.  
\colorbox{gray!20}{\strut Diverse errors}, 
\colorbox{green!20}{\strut ANLI only tuning}, 
\colorbox{blue!20}{\strut ANLI + synthetic mix} indicate the type of row. 
}
\label{tab:results}
\end{table}

\subsection{RQ2: Error types and synthesis logic}

\noindent \textbf{Effect of Error Types.}  
Figure~\ref{fig:error_types} shows that, across benchmarks and models, incorporating diverse error types generally improves verification scores compared to using only unverifiable errors, with the sole exception of \texttt{SynClaimEval} on \texttt{Qwen}.  
This underscores the value of error-type variation during tuning for enhancing model robustness.  
   
\noindent \textbf{Complexity of claims}
Table \ref{tab:results} shows that introducing structure into synthesis further shapes model behavior.  
For \texttt{LLaMA}, structured variants outperform unstructured ones: context-graph synthesis yields moderate improvements, while argument-graph synthesis delivers the strongest results, atleast at lower context sizes. This ordering---\emph{argument-graph $>$ context-graph $>$ unstructured}---highlights the benefit of conditioning on richer discourse and argumentative structure. 
In contrast, \texttt{Qwen} again shows limited variation across synthesis strategies, suggesting that structural supervision is more valuable for weaker models that lack strong baseline verification ability.

\noindent \textbf{Mixing Synthesis Strategies.}  
We evaluate strategy mixing on \texttt{LLaMA}, the model that benefited most from synthesis.  
Table~\ref{tab:results} shows that combining strategies yields higher performance than any single strategy, particularly on \texttt{FinDver} (0.63 F1) and \texttt{SynClaimEval} (0.81), while \texttt{UniSummEval} shows a slight drop.  
We hypothesize that this decline reflects differences in average context length, as both \texttt{SynClaimEval} and \texttt{FinDver} consist of longer inputs.  

\noindent \textbf{Using Synthesis for Augmentation.}  
Table \ref{tab:results} , last $3$ rows, shows that augmenting the mixed strategy with \texttt{ANLI} yields the strongest overall results, reaching 0.82 F1 on \texttt{SynClaimEval} and 0.65 on \texttt{FinDver}.  
These scores surpass tuning with \texttt{ANLI} or synthetic data alone, underscoring the benefits of synthetic claims as complementary augmentation.  


\subsection{RQ3 Impact on generated explanations}

We apply the ranking formula from \S\ref{subsec:explain_eval} to all synthesis variants. 
Figure~\ref{fig:expl_ranking_composite} shows that for \texttt{LLaMa}, a consistent ordering emerges across all four benchmarks: \emph{argument-graph} $>$ \emph{context-graph} $>$ \emph{unstructured} $>$ \emph{base model}. 
The highest ranking scores are obtained by the \emph{argument-graph} variants with $16k$ context length, followed by context-graph based synthesis, while unstructured synthesis trails behind. 
This ordering mirrors our quantitative results, reinforcing the finding that structured synthesis---particularly when applied with longer contexts---is more beneficial than either unstructured synthesis or no finetuning
\footnote{Illustrative examples of generated rationales are provided in Appendix~\ref{app:reasoning_out}.} .
By contrast, the trends for \texttt{Qwen} differ. 
Here, only argument-graph synthesis yields clear improvements over the base model, while context-graph synthesis shows limited gains and unstructured synthesis consistently ranks lowest. 
This divergence suggests that while synthetic tuning can enhance both prediction scores and explanation quality, its impact depends strongly on the underlying model family. 
Taken together, these findings highlight both the promise and the limitations of synthetic data: structured synthesis can promote more supportive rationales, but its benefits are not uniformly transferable across architectures.


\section{Conclusion and Future Work}
We introduced \texttt{SynClaimEval}, a framework for evaluating the utility of synthetic data in long-context claim verification. By disentangling three dimensions—context length, synthesis logic, and explanation quality—we found that synthetic fine-tuning can improve verification accuracy, particularly under structured synthesis settings that expose models to more complex claims, though these gains are not always consistent. Beyond accuracy, synthetic data proves valuable as an augmentation to human-written claims and more reliably enhances explanation quality, especially with argument-graph synthesis. Looking forward, applying \texttt{SynClaimEval} to more diverse and domain-specific settings, and combining synthetic with human-annotated data, will be key to understanding the broader impact of synthetic training on long-context reasoning.




\section*{Limitations}
Our study evaluated several long-context synthesis strategies for claim verification, but important limitations remain. First, we relied on widely available public datasets as synthesis sources. While this choice ensures reproducibility, it also risks overlap with model pretraining corpora. Future work should incorporate more diverse and domain-specific sources to better probe generalization and reduce contamination effects. Second, we restricted training to supervised fine-tuning (SFT). Exploring alternative paradigms—such as reinforcement learning or domain-adaptive pretraining—could reveal different trade-offs between generalization and explanation quality. Third, we limited our experiments to parameter-efficient tuning; extending the framework to full-parameter tuning may yield additional insights. Fourth, scaling synthesis to more challenging domains (e.g., scientific, legal, or financial texts where LLMs often struggle) would clarify how task complexity mediates the benefits of synthetic data. Finally, our explanation-quality assessment relied on LLM-based judges, which, while cost-effective, may introduce biases. Complementing them with human evaluation remains an important direction.


\section*{Ethics Statement}
This work relies exclusively on publicly available datasets for both synthesis and evaluation, which minimizes risks of handling sensitive or private information. Nevertheless, synthetic data generation may inadvertently amplify biases present in the underlying sources or in the language models used for synthesis. We attempt to mitigate this by sampling from diverse domains and by analyzing multiple synthesis strategies, but acknowledge that residual bias may remain. 

\section*{Acknowledgment}
I would like to sincerely thank my manager during the internship, Matthew Danielson, for his mentorship and steady feedback throughout this work. I am grateful to Amir Rez Rahmani, Bin He, and Winston Quock for their regular feedback and thorough reviews. I also thank the Legal team—especially Abigail Holman—for guidance through the publication process, and the Human-in-the-Loop (HITL) team at Zillow for annotation work and careful reviews. Finally, I thank the LLM Platform (LLMP) team and Taleb Zeghmi for engineering support and access to GPU resources that enabled the experiments.

\bibliography{anthology,custom}
\bibliographystyle{acl_natbib}

\appendix





\section{Summarization Prompts}
\label{app:summ_prompts}
Table~\ref{tab:sum-prompts} presents the domain-specific summarization prompts used to compress inputs from various domains to generate synthetic data.  
Each template is tailored to the conventions of its source domain (e.g., government reports, meeting transcripts, scientific articles, or books), while enforcing common constraints such as conciseness, professional tone, and length limits.


\begin{table*}[t]
\small
\centering
\begin{tabularx}{\textwidth}{@{}l X@{}}
\toprule
\textbf{Domain} & \textbf{Prompt Template} \\
\midrule

\rowcolor{gray!5}
GovReports &
\begin{minipage}[t]{\linewidth}\ttfamily
Your task is to write a concise, structured summary for the government report below.

Organize your summary into multiple paragraphs. Use a clear, professional tone. Keep the total length under 1000 words. Do not include the full report title in your summary---refer to it generically as ``the report.''

Report\\
\{input\_text\}

Summary:
\end{minipage}
\\
\addlinespace[8pt]

MeetingBank &
\begin{minipage}[t]{\linewidth}\ttfamily
Your task is to produce a concise, structured ``mini'' summary of the meeting transcript below (e.g., as in MeetingBank). Treat the summary as a compact representation that captures all essential discussion points and outcomes.

Additional requirements:\\
- Keep the summary under 1000 words.\\
- Do \textbf{not} include verbatim transcript excerpts---paraphrase in your own words.\\
- Use consistent terminology (e.g., refer to ``Project X'' the same way throughout).

Transcript\\
\{input\_text\}

Summary:
\end{minipage}
\\
\addlinespace[8pt]

\rowcolor{gray!5}
PubMed &
\begin{minipage}[t]{\linewidth}\ttfamily
Your task is to write a concise, structured ``mini'' version of the scientific document below. Treat the summary as a compact version of the input that retains all critical content.

Additional requirements:\\
- Organize the summary into multiple paragraphs.\\
- Use full technical names on first mention, then acronyms thereafter.\\
- Keep the summary under 1000 words.\\
- Do \textbf{not} include the document's title or citation details---focus only on content.\\
- Ensure the summary reads as a true ``mini'' of the input, condensing its essence into a coherent, readable format.

Document\\
\{input\_text\}

Summary:
\end{minipage}
\\
\addlinespace[8pt]

SQuALITY / Books &
\begin{minipage}[t]{\linewidth}\ttfamily
Your task is to write a summary for the book below. Include vital information about key events, backgrounds, settings, characters, their objectives, and motivations. Introduce characters (with full names), places, and other major elements on first mention. The book may feature non-linear narratives (flashbacks, alternate worlds/viewpoints). Organize the summary into a consistent, chronological narrative. The summary must be under 1000 words, span multiple paragraphs, and be written as a single continuous narrative (no bullet lists or outlines). Do \textbf{not} include the book name in the summary.

Book\\
\{input\_text\}

Summary:
\end{minipage}
\\

\bottomrule
\end{tabularx}
\caption{\centering Summarization prompt templates used for synthetic data generation across four domains. Each template specifies domain-specific constraints and formatting requirements, while maintaining consistency in output length and style. Replace \texttt{\{input\_text\}} with the source document.}
\label{tab:sum-prompts}
\end{table*}

\section{Claim Synthesis Prompts}
\label{app:generation_prompts}

\subsection{Unstructured Synthesis}
Table~\ref{tab:claim-unstructured} presents the prompts used to generate verifiable, unverifiable (hallucination-based), and contradictory claims.  
To ensure a strict 1:1 mapping across verification types, we first synthesize verifiable claims and then apply corruption procedures to derive their unverifiable and contradictory counterparts.  

\begin{table*}[t]
\small
\centering
\begin{tabularx}{\textwidth}{@{}l X@{}}
\toprule
\textbf{Synthesis Type} & \textbf{Prompt Template} \\
\midrule

\rowcolor{blue!5}
Verified &
\begin{minipage}[t]{\linewidth}\ttfamily
You are given a document. Your task is to extract a list of \{num\_claims\} factual claims from the document.

Each claim must:
- Be a complete, standalone statement that can be independently verified.  
- Be factual, atomic, clear, and concise.  
- Be grounded in the document (no hallucinations).  
- Be diverse (avoid closely related claims).  

For each claim, provide reasoning showing why it is factual and supported.

Return only the following format:  

<BEGINFACT>Factual statement<ENDFACT>  
<BEGINREASONING>Explanation<ENDREASONING>

Document: \{input\}
\end{minipage}
\\
\addlinespace[6pt]

\rowcolor{yellow!10}
Unverifiable &
\begin{minipage}[t]{\linewidth}\ttfamily
You are given a factual claim from a document. Generate a \textbf{plausible but unverifiable variant}.  

It must:  
- Sound realistic and grammatically correct.  
- Be related to the topic but include unverifiable information.  
- Not be explicitly contradictory.  

Output only:  

<BEGINUNVERIFIABLE>Unverifiable claim<ENDUNVERIFIABLE>  
<BEGINUNVERIFIABLEREASON>Reason why unverifiable<ENDUNVERIFIABLEREASON>

Document: \{document\}  
Claim: \{factual\_claim\}
\end{minipage}
\\
\addlinespace[6pt]

\rowcolor{red!10}
Contradictory &
\begin{minipage}[t]{\linewidth}\ttfamily
You are given a factual claim. Generate a \textbf{corrupted version} using a specific error type: \{error\_type\}.  

Error types:  
- negation (flip polarity)  
- entity\_relation (swap/alter entities or relations)  
- discourse (flip cause–effect or misattribute support)  

If not feasible, return <NOT\_POSSIBLE>.  

Output only:  

<BEGINFALSIFIED>Falsified claim<ENDFALSIFIED>  
<BEGINFALSEREASON>Reasoning<ENDFALSEREASON>  
<BEGINERRORTYPE>\{error\_type\}<ENDERRORTYPE>

Document: \{document\}  
Factual Claim: \{factual\_claim\}
\end{minipage}
\\

\bottomrule
\end{tabularx}
\caption{\centering Unstructured claim synthesis prompts. Each synthesis type is shaded for clarity: \colorbox{blue!10}{Verified}, \colorbox{yellow!10}{Unverifiable}, and \colorbox{red!10}{ Contradictory}. Placeholders \{\} are replaced with inputs during generation.}
\label{tab:claim-unstructured}
\end{table*}

\subsection{Context-graph Synthesis Prompts}
Table~\ref{tab:doc2ent} presents the prompt used to extract entity triplets from the input document.  
Building on these outputs, Table~\ref{tab:claim-context} provides the synthesis prompts for generating verifiable, unverifiable, and contradictory claims, each of which consumes the extracted entities as input.  

\begin{table*}[t]
\small
\centering
\begin{tabularx}{\textwidth}{@{}X@{}}
\toprule
\textbf{Document $\rightarrow$ Entity Triples Extraction Prompt} \\
\midrule
\begin{minipage}[t]{\linewidth}\ttfamily
Given an article, go over every sentence and extract triples in the form:  
(entity <TUPLEDELIM> entity <TUPLEDELIM> short description of the relation).  

Group triples with the same entity together.  
Separate groups using <GROUPDELIM>.  

Provided Sentences:  
\{input\}  

Groups of Triples in Provided Document:
\end{minipage} \\
\bottomrule
\end{tabularx}
\caption{Prompt for extracting entity–entity–relation triples from a document (\textbf{Document $\rightarrow$ Entities} step).}
\label{tab:doc2ent}
\end{table*}

\begin{table*}[t]
\small
\centering
\begin{tabularx}{\textwidth}{@{}l X@{}}
\toprule
\textbf{Context-Graph Synthesis Type} & \textbf{Prompt Template} \\
\midrule

\rowcolor{blue!5}
\textbf{Verified (uses given entities)} &
\begin{minipage}[t]{\linewidth}\ttfamily
You are given a document. Write a \textbf{single factual claim} that \textbf{must mention all of the following entities}:

\textbf{Entities}: \{entities\}

Then provide a \textbf{brief explanation} grounded in the document.

Output exactly:

<BEGINFACT>Your factual claim using all entities.<ENDFACT> \\
<BEGINREASONING>Why the claim is factual and supported by the document.<ENDREASONING>

Document: \{input\}
\end{minipage}
\\
\addlinespace[6pt]

\rowcolor{yellow!10}
\textbf{Unverifiable Variant (same entities)} &
\begin{minipage}[t]{\linewidth}\ttfamily
You are given a factual claim involving the entities \{entities\}. Generate a \textbf{plausible but unverifiable} variant that \textbf{introduces at least one relationship not verifiable} from the document (avoid explicit contradiction).

Output exactly:

<BEGINUNVERIFIABLE>Unverifiable claim with the same entities.<ENDUNVERIFIABLE> \\
<BEGINUNVERIFIABLEREASON>This claim ... (explain why unverifiable without referencing the original claim).<ENDUNVERIFIABLEREASON>

\textbf{Document:} \{document\} \\
\textbf{Claim:} \{factual\_claim\} \\
\textbf{Entities:} \{entities\}
\end{minipage}
\\
\addlinespace[6pt]

\rowcolor{red!10}
\textbf{Contradictory Variant (same entities)} &
\begin{minipage}[t]{\linewidth}\ttfamily
You are given a factual claim involving the entities \{entities\}. Generate a \textbf{contradictory} variant by \textbf{flipping or corrupting at least one relationship} among these entities (keep entities unchanged). The new claim must be \textbf{contradicted} by the document (not merely unverifiable).

Output exactly:

<BEGINFALSIFIED>Contradictory claim with the same entities.<ENDFALSIFIED> \\
<BEGINFALSEREASON>This claim ... (explain why contradicted, citing the corrupted relationship).<ENDFALSEREASON>

\textbf{Document:} \{document\} \\
\textbf{Claim:} \{factual\_claim\} \\
\textbf{Entities:} \{entities\}
\end{minipage}
\\

\bottomrule
\end{tabularx}
\caption{\centering Context-graph (structured) claim prompts. Row colors indicate type: \colorbox{blue!10}{Verified}, \colorbox{yellow!10}{Unverifiable}, and \colorbox{red!10}{Contradictory}. The triple-extraction step is omitted here for space; this table assumes entities are already provided.}
\label{tab:claim-context}
\end{table*}

\subsection{Argument-graph Synthesis Prompts}
Table~\ref{tab:doc2arg} shows the prompt for extracting argument roles—\texttt{claims} and \texttt{premises}—along with their support/oppose relations.  
These roles are assembled into an argument graph, from which connected chains are sampled and passed to the synthesis prompts in Table~\ref{tab:arg-claim-prompts}.   
\begin{table*}[t]
\small
\centering
\begin{tabularx}{\textwidth}{@{}X@{}}
\toprule
\textbf{Argument Graph Extraction Prompt (Document $\rightarrow$ Argument Graph)} \\
\midrule
\begin{minipage}[t]{\linewidth}\ttfamily
Given a passage, extract its argument structure by identifying \textbf{claims}, \textbf{premises}, and the \textbf{relation} between each premise and its claim (\emph{supports} or \emph{opposes}).

A \textbf{claim} is the main assertion.  
A \textbf{premise} is a reason/evidence that supports or opposes the claim.

For each claim, list all connected premises with their relation.

\#\#\# Output Format (repeat per group):
<BEGIN\_GROUP\_CLAIM>  
<STARTCLAIM>The claim goes here<ENDCLAIM>  
<STARTPREMISE>Premise text<STARTRELATION>supports or opposes<ENDRELATION><ENDPREMISE>  
... (repeat premise blocks as needed)  
<END\_GROUP\_CLAIM>

Only include relations explicitly inferable from the passage.  
Do not include general facts, summaries, or hallucinated reasoning.

Input:  
\{input\_text\}
\end{minipage}
\\
\bottomrule
\end{tabularx}
\caption{\centering Prompt for constructing an \textbf{argument graph} from a document (claims, premises, and support/oppose links).}
\label{tab:doc2arg}
\end{table*}

\begin{table*}[t]
\small
\centering
\begin{tabularx}{\textwidth}{@{}l X@{}}
\toprule
\textbf{Argument-Graph Synthesis Type} & \textbf{Prompt Template} \\
\midrule

\rowcolor{blue!5}
\textbf{Verified (from argument chain)} &
\begin{minipage}[t]{\linewidth}\ttfamily
Given an \textbf{argument chain} (a central claim with connected premises and their relations: \emph{supports}/\emph{opposes}) and the reference document, generate \textbf{one concise, overarching factual claim} that synthesizes the core argument. Integrate both supporting and opposing premises faithfully.

Provide a brief, document-grounded explanation.

Output exactly:
<BEGINFACT>Your factual claim synthesizing the chain.<ENDFACT>  
<BEGINREASONING>Why the claim is factual, grounded in the document.<ENDREASONING>

Document: \{input\}  
Argument Chain: \{argument\_chain\}
\end{minipage}
\\
\addlinespace[6pt]

\rowcolor{yellow!10}
\textbf{Unverifiable (from argument chain)} &
\begin{minipage}[t]{\linewidth}\ttfamily
Given an \textbf{argument chain} and the reference document, generate \textbf{one plausible claim} that integrates the chain but introduces an \textbf{unverifiable detail} (cannot be confirmed from the document; avoid contradiction).

Then explain why it is unverifiable (identify the unconfirmed part). Start reasoning with “This claim...”.

Output exactly:
<BEGINUNVERIFIABLE>Your unverifiable, chain-based claim.<ENDUNVERIFIABLE>  

<BEGINUNVERIFIABLEREASON>This claim ... (why unverifiable, based on what is missing/uncertain in the document).<ENDUNVERIFIABLEREASON>

Document: \{document\}  
Argument Chain: \{argument\_chain\}
\end{minipage}
\\
\addlinespace[6pt]

\rowcolor{red!10}
\textbf{Contradictory (flip relation in chain)} &
\begin{minipage}[t]{\linewidth}\ttfamily
Given an \textbf{argument chain} and the reference document, generate \textbf{one concise claim} that \textbf{falsifies the original argument} by \textbf{incorrectly flipping} at least one premise relation (treat a supporting premise as \emph{opposes}, or vice versa). The result must be \textbf{contradicted} by the document (not merely unverifiable).

Then explain why it is falsified, citing the misrepresented relationship.

Output exactly:
<BEGINFALSIFIED>Your falsified claim that flips a support/oppose relation.<ENDFALSIFIED>  
<BEGINFALSEREASON>Why this claim is contradicted (what relation was flipped and how the document disagrees).<ENDFALSEREASON>

Document: \{document\}  
Argument Chain: \{argument\_chain\}
\end{minipage}
\\

\bottomrule
\end{tabularx}
\caption{Argument-graph (structured) claim prompts spanning two columns. Row colors indicate type: \colorbox{blue!10}{Verified}, \colorbox{yellow!10}{Unverifiable}, \colorbox{red!10}{Contradictory}. This table assumes the argument graph has been extracted using Table~\ref{tab:doc2arg}.}
\label{tab:arg-claim-prompts}
\end{table*}

\section{Error types definitions}
\label{app:err_types}
Table~\ref{tab:error-types} outlines the error granularities considered when synthesizing unverified claims. 

\begin{table*}[ht!]
\small
\centering
\begin{tabularx}{\columnwidth}{@{}l X@{}}
\toprule
\textbf{Error Type} & \textbf{Definition / Transformation Strategy} \\
\midrule

\rowcolor{gray!5}
\textbf{Unverifiable} & Produce a claim that sounds plausible but cannot be verified from the source (e.g., by introducing unverifiable details while avoiding explicit contradiction). \\

\textcolor{red}{\textbf{Negation}} & \textcolor{red}{Flip the polarity of the claim to create a false statement (e.g., ``X occurred'' $\rightarrow$ ``X did not occur'').} \\

\rowcolor{gray!5}
\textcolor{red}{\textbf{Entity-Relation}} & \textcolor{red}{Corrupt entities or their relationships, such as swapping subject/object roles, misattributing actions, or replacing entities with plausible but incorrect ones.} \\

\textcolor{red}{\textbf{Discourse}} & \textcolor{red}{Corrupt the logical structure of the claim, e.g., flipping cause–effect, reversing claim and evidence, or misrepresenting support/oppose relations.} \\

\bottomrule
\end{tabularx}
\caption{\centering Error types used in synthetic claim generation. \textcolor{red}{Red rows} denote contradictory error types, while unverifiable errors add uncertainty without explicit contradiction.}
\label{tab:error-types}
\end{table*}

\section{GPT-4o Evaluation of Claim Synthesis}
\label{app:auto_claim_eval}
Table ~\ref{tab:gpt-4o-synthetic-claims} captures the quality of synthetic claims across different dataset and context length. We pass the generated claim along with relevant document and leverage GPT-4o as a judge to understand the quality of generated data measured in terms of accuracy

\begin{table*}[t]
\small
\centering
\begin{tabular}{lcccccc}
\toprule
\textbf{Dataset} & \textbf{Length} & \textbf{No Error} & \textbf{Unverifiable} & \textbf{Negation} & \textbf{Entity Rel.} & \textbf{Discourse} \\
\midrule
GovReport   & 4k  & 0.88 & 0.86 & 0.98 & 0.82 & 0.86 \\
            & 16k & 1.00 & 0.92 & 1.00 & 0.80 & 0.72 \\
\addlinespace
SQuALITY    & 4k  & 0.92 & 0.94 & 0.96 & 0.88 & 0.86 \\
            & 16k & 0.96 & 0.92 & 1.00 & 0.86 & 0.78 \\
\addlinespace
MeetingBank & 4k  & 0.96 & 0.80 & 0.98 & 0.76 & 0.80 \\
            & 16k & 0.92 & 1.00 & 1.00 & 0.84 & 0.76 \\
\addlinespace
PubMed      & 4k  & 0.96 & 0.90 & 1.00 & 0.80 & 0.76 \\
            & 16k & 1.00 & 0.94 & 0.98 & 0.96 & 0.84 \\
\bottomrule
\end{tabular}
\caption{\centering GPT-4o evaluation accuracy of synthetic claims under 4k vs 16k unstructured settings, reported per dataset and error type.}
\label{tab:gpt-4o-synthetic-claims}
\end{table*}

\section{Evaluating the quality of synthetic explanations}
\label{app:exp_quality}
\noindent \textbf{Quality of generated explanations}
Following \cite{pham2025clipper} which evaluated informativeness/faithfulness of the CoT through grounding each step to the input, we evaluate how well generated explanations remain grounded before and after synthesis. We decompose each explanation into atomic facts with \texttt{GPT-4.1}, and we compute the proportion of those facts that can be verified against the original context across all synthetic strategies. We sample $100$ generated explanations from each synthetic strategy from the verifiable label.  At the $4$k truncation level, unstructured synthesis achieved $86.12\%$ verified units, context-graph synthesis achieved $80.72\%$, while argument-graph synthesis attained the highest verification rate at $93.57\%$. At the $16$k truncation level, unstructured ($89.39\%$) and context-graph ($88.32\%$) synthesis improved compared to their $4$k counterparts, though argument-graph synthesis remained strong ($91.11\%$). These numbers are in the same range with prior findings of synthetic  CoT faithfulness described in \cite{pham2025clipper}, which showed benefits of synthetic claim generation.

Table ~\ref{tab:split-reasoning-prompt} shows the prompts for extracting atomic claims from model generated reasoning justifying the final judgment. Once the atomic claims are extracted Table ~\ref{tab:fact-eval-prompt} shows the prompts used to evaluate the correctness of the atomic fact and finally evaluated the quality of CoT reasoning used for training the models

\begin{table*}[t]
\small
\centering
\begin{tabularx}{\textwidth}{@{}l Y@{}}
\toprule
\textbf{Prompt} & \textbf{Content} \\
\midrule
Atomic Fact Extraction (Split Reasoning) &
\begin{minipage}[t]{\linewidth}\ttfamily
\#\# \textbf{Task Description}\\
You will be given an explanation statement. Your task is to extract a set of \textbf{atomic facts}---statements that can be \textbf{directly inferred} from this explanation without interpretation, additional assumptions, or redundancy.\\[4pt]

\#\# \textbf{Guidelines:}\\
- Extract only \textbf{explicitly stated} atomic facts in the explanations.\\
- \textbf{Do not repeat} facts or include any that require external knowledge.\\
- Maintain \textbf{granularity}: Each fact should be \textbf{minimal yet complete}.\\
- Structure your output as a valid list of facts, \textbf{one fact per line}. \textbf{Do not include any additional text or formatting.}\\
- Each summary has at least 1 atomic fact.\\[4pt]

\---\\
\#\# \textbf{Example Output Format}\\
"First atomic fact"\\
"Second atomic fact"\\
"Third atomic fact"\\
\---\\[4pt]

\#\# \textbf{Input}\\
\textbf{Explanation:}\\
\{explanation\}\\[4pt]

\---\\
\#\# \textbf{Output}\\
(\textbf{List Only})
\end{minipage}
\\
\bottomrule
\end{tabularx}
\caption{\centering Split-reasoning prompt for extracting atomic facts from an explanation. Replace \texttt{\{explanation\}} with the input text.}
\label{tab:split-reasoning-prompt}
\end{table*}

\begin{table*}[t]
\small
\centering
\begin{tabularx}{\textwidth}{@{}l Y@{}}
\toprule
\textbf{Prompt} & \textbf{Content} \\
\midrule
Atomic Fact Support Evaluation (yes/no) &
\begin{minipage}[t]{\linewidth}\ttfamily
\#\# \textbf{Task Description}\\
You are given an \textbf{atomic fact} and a \textbf{context}.\\
Your task is to determine whether the fact is \textbf{fully supported} by the context.\\[4pt]

\#\# \textbf{Guidelines:}\\
- A fact is \textbf{supported} only if all of its information is explicitly confirmed by the context.\\
- If any part of the fact is missing, contradicted, or not stated in the context, mark it as \textbf{not supported}.\\
- Respond with only: \textbf{yes} (fully supported) or \textbf{no} (not fully supported).\\[4pt]

\---\\
\#\# \textbf{Input}\\
\textbf{Context:}\\
\{context\}\\[2pt]
\textbf{Fact:}\\
\{fact\}\\[4pt]

\---\\
\#\# \textbf{Output (yes/no only)}
\end{minipage}
\\
\bottomrule
\end{tabularx}
\caption{Prompt for evaluating whether an atomic fact is fully supported by a given context. Replace \texttt{\{context\}} and \texttt{\{fact\}} with inputs.}
\label{tab:fact-eval-prompt}
\end{table*}

\begin{table*}[t]
\small
\centering
\begin{tabularx}{\linewidth}{@{}l Y@{}}
\toprule
\textbf{Prompt} & \textbf{Content} \\
\midrule
Reasoning Ranking &
\begin{minipage}[t]{\linewidth}\ttfamily
The true label is '\{label\}'. Both explanations aim to justify this decision.\\[6pt]
1) \{r1\}\\
2) \{r2\}\\[6pt]
Which explanation is more detailed and supportive? Reply with '1', '2', or 'equal'.
\end{minipage}
\\
\bottomrule
\end{tabularx}
\caption{\centering Prompt for pairwise reasoning ranking: given a gold label and two competing explanations, the judge selects which explanation is more detailed and supportive.}
\label{tab:reasoning-ranking}
\end{table*}

\label{app:llm_judge_ranking}


\label{app:eval_synthetic_data}

\section{Reasoning Output}
\label{app:reasoning_out}
Table ~\ref{tab:reasoning-comparison} shows the comparison of model-generated explanation under different synthesis strategies and help understand the impact complex synthesis strategies like Argument-Graph has on model-generated explanations.
\begin{table*}[t]
\small
\centering
\begin{tabularx}{\textwidth}{@{}l Y@{}}
\toprule
\textbf{Model / Setting} & \textbf{Generated Reasoning (Label = No)} \\
\midrule

Base Model &
The claim is inconsistent with the document. \\

\addlinespace[4pt]
Unstructured &
The claim is inconsistent with the document. The document states that OSG is exploring opportunities in transporting liquified CO\textsubscript{2}. \\

\addlinespace[4pt]
Context-Graph &
The claim is inconsistent with the document. The document states that OSG is exploring opportunities in transporting liquified CO\textsubscript{2}, which is relevant to maritime innovations and aligns with the company's strategy to participate in emerging markets for energy and liquid bulk commodities. However, the claim incorrectly suggests that all revenue from maritime innovations is allocated to sectors outside the maritime industry, which contradicts the document's statement that OSG is actively pursuing such opportunities. \\

\addlinespace[4pt]
Argument-Graph &
The claim is inconsistent with the document. The document states that OSG is exploring opportunities in transporting liquified CO\textsubscript{2}, which is a natural next step into an emerging market, consistent with OSG's expertise with liquid cargoes. However, it does not mention that all revenue is allocated to sectors outside maritime innovations. The document highlights that OSG is seeking opportunities to increase scale and drive cost efficiencies through a disciplined approach to investment in core and adjacent asset classes to maximize return on capital across market cycles. This includes the recent acquisition of the Alaskan Frontier and the planned life cycle upgrades on the four Alaska class vessels, which are expected to improve performance and fuel efficiency, leading to a meaningful reduction in carbon output. Additionally, the document mentions that OSG is committed to acting in the best interests of its stockholders and is carefully considering Saltchuk Resources' indication of interest to acquire all outstanding shares of OSG's common stock that Saltchuk does not own for \$6.25 per share in cash. The document does not provide any information about revenue allocation outside the maritime sector. \\

\bottomrule
\end{tabularx}
\caption{\centering Comparison of model-generated explanations under different synthesis strategies (label = \texttt{No}). Structured synthesis produces longer, more detailed rationales that incorporate specific evidence.}
\label{tab:reasoning-comparison}
\end{table*}

\end{document}